\documentclass[journal]{IEEEtran}
\IEEEoverridecommandlockouts
\usepackage{mathrsfs}
\usepackage{amssymb}
\usepackage{float}
\usepackage{color}
\definecolor{shadecolor}{rgb}{1,1,0}
\usepackage{framed}
\usepackage{amsmath}
\usepackage{amssymb}
\usepackage{amsthm}
\usepackage{multirow,makecell}
\usepackage{setspace}
\usepackage{stfloats}
\usepackage{graphicx}
\usepackage{epstopdf}
\usepackage{soul}
\usepackage{colortbl}
\usepackage[table ]{ xcolor}
\soulregister\cite7
\soulregister\ref7
\usepackage[textsize=small]{todonotes}
\ifCLASSINFOpdf \else \fi

\usepackage{algorithm}
\usepackage{algorithmic}

\usepackage{cite}

\usepackage[normalem]{ulem}
\usepackage{soul}
\usepackage{ulem}

\begin{document}
\title{The Performance Analysis of Spectrum Sharing between UAV enabled Wireless Mesh Networks and Ground Networks}
\author{Zhiqing Wei,
        Jialin Zhu,
        Zijun Guo,
        and Fan Ning
\thanks{An earlier version of this paper was presented at the WCSP 2018 Conference and was published in its Proceedings \cite{wcsp_hangzhou}.
The URL of this conference paper is: https://ieeexplore.ieee.org/document/8555855. This work was supported in part by the Beijing Natural Science Foundation (No. L192031) and the National Natural Science Foundation of China under Grant 61631003.

Zhiqing Wei, Jialin Zhu, Zijun Guo, and Fan Ning
are with Key Laboratory of Universal Wireless Communications, Ministry of Education, Beijing University of Posts and Telecommunications, Beijing 100876, China (email: \{weizhiqing, jialinzhu, zijunguo, ningfan\}@bupt.edu.cn).}}

\maketitle

\begin{abstract}
Unmanned aerial vehicle (UAV)
has the advantages of large coverage and flexibility,
which could be applied in
disaster management to provide wireless services
to the rescuers and victims.
When UAVs forms an aerial mesh network,
line-of-sight (LoS) air-to-air (A2A)
communications have long transmission distance,
which extends the coverage of multiple UAVs.
However, the capacity of
UAV is constrained due to the multiple hop transmissions
in aerial mesh networks. In this paper, spectrum sharing between UAV enabled wireless mesh
	networks and ground networks is studied to improve
	the capacity of UAV networks. Considering
	two-dimensional (2D) and three-dimensional (3D)
	homogeneous Poisson point process (PPP)
	modeling for the distribution of UAVs
	within a vertical range $\Delta h$,
	stochastic geometry is applied to analyze
	the impact of the height of UAVs,
	the transmit power of UAVs,
	the density of UAVs and
	the vertical range, etc., on
	the coverage probability of ground network user
	and UAV network user.
Besides, performance improvement of spectrum sharing
with directional antenna is verified.
With the object function of
maximizing the transmission capacity,
the optimal altitude of UAVs
is obtained.
This paper provides a theoretical
	guideline for the spectrum sharing
	of UAV enabled wireless mesh networks,
	which may contribute significant value to the study of spectrum sharing mechanisms
	for UAV enabled wireless mesh networks.
\end{abstract}

\IEEEpeerreviewmaketitle

\section{Introduction}

UAV enabled wireless communication has unique advantages
of long signal propagation distance and flexible deployment \cite{UAV_Survey0},
which has attracted wide
attention.
Among the applications of UAVs,
UAV-mounted BSs can establish wireless connections for
mobile devices in disaster management \cite{UAV_Multi}
and offload traffic from
ground base stations (BSs) with heavy traffic load \cite{wu}.

\begin{figure}[!t]
	\centering
	\includegraphics[width=0.5\textwidth]{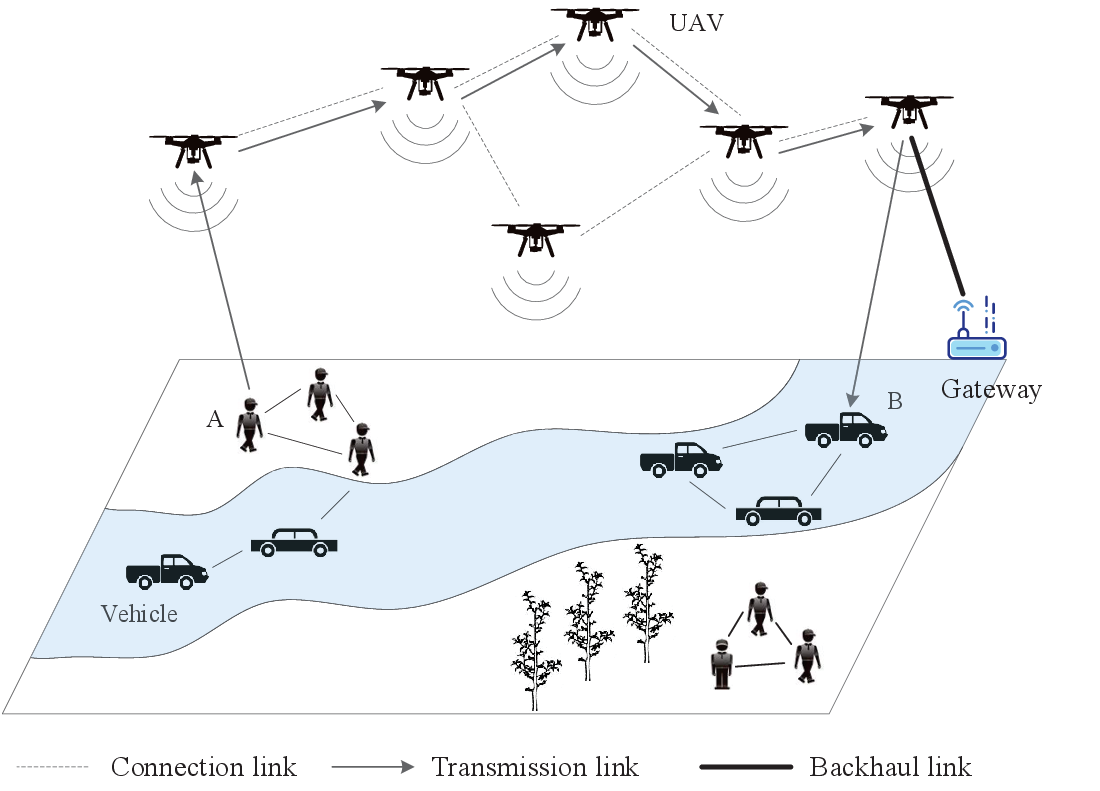}
	\caption{The application scenario of UAV mesh network.}
	\label{Figure_1}
\end{figure}

Multiple UAVs can provide
wireless services efficiently \cite{UAV_Multi}.
Chand \emph{et al.} \cite{UAV_MESH_1}
proposed a framework of UAV mesh network
to support disaster management and military operations.
Lyu \emph{et al.} \cite{UAV_Deployment_Geo}
designed a geometrical UAV placement method
to serve the ground users with minimum number of UAVs.
Bor-Yaliniz \emph{et al.} \cite{UAV_Deployment_Environment}
designed a UAV placement method considering environment parameters
such as shapes of buildings.
Plachy \emph{et al.} \cite{UAV_positioning_association}
jointly optimized placement of UAV-mounted BSs and
user association scheme.
Lu \emph{et al.} \cite{UAV_lu}
designed a measurement-aided dynamic planning
to optimize placement and channel allocation
of UAV-mounted BSs.
In \cite{ICCC}, a UAV mesh network is developed by our team.
The scenario of \cite{ICCC} is illustrated in Fig. \ref{Figure_1},
where the UAV mesh network provides
connections for ground users.

\begin{table*}[!t]
	\caption{Key Parameters and Notations}
	\label{tab_1}
	\begin{center}
		\begin{tabular}{l l}
		\hline
		\hline
		{Symbol} & {Description} \\	
		$P_u$ & Transmit power of UAV\\
		$P_d$ & Transmit power of TXs of ground network\\
		${\lambda_d}$ & TXs density of ground network\\
		${\lambda_{2Du}}$ & User density of 2D UAV network\\
		${\lambda_{3Du}}$ & User density of 3D UAV network\\
		$h_1$ & Minimum height of 3D UAV network \\
		$\Delta h$ & Vertical range of 3D UAV network\\
		$\alpha_d$ & Path-loss exponent of ground-to-ground link\\
		$\alpha_u$ & Path-loss exponent of air-to-ground link\\
		$N$ & Variance of additive white Gaussian noise\\
		$\eta$ & Attenuation factor due to NLoS propagation\\
		$\beta$ & SINR threshold\\
		$B$ and $C$ & Environmental dependent constants\\
		$\theta$ &  Elevation angle of air-to-ground link by line-of-sight transmission\\
		${\gamma_{gu}}$ & Received SINR of typical ground network user\\
		$\gamma_{uu}$ & Received SINR of UAV network user\\
		${I_{gu}^c}$ & Interference generated by TXs of ground network\\
		${{I_u}}$ & Interference generated by TXs of UAV network\\
		${I_u^c}$ & Received interference from TXs of UAV network\\
		${P_1}$ & Coverage probability of typical ground network user\\
		${P_2}$ & Coverage probability of typical UAV network user\\
		${L_A}(*)$ & Laplace transform of random variable $A$\\
		$g_0$ & Small scale fading gain between typical ground network user and its associated TX\\
		$g_i$ & Small scale fading gain of $i$th interference link\\
		$d_0$ & Distance between typical ground network user and its associated TX\\
		$d_i$ & Distance between $i$th BS of ground network and	the user at the origin $\{ \bf{0}\}$\\
		$x_0$ & Distance between typical UAV network user and its associated UAV\\
		$x_i$ & Distance between $i$th UAV and the user at the origin $\{ \bf{0}\}$\\	
		\hline
		\hline
	\end{tabular}
	\end{center}
\end{table*}

According to \cite{Kumar}, as the number of nodes increases, the capacity of wireless mesh networks decreases dramatically because of the existence of the multiple hop transmissions.
	As illustrated in Fig. \ref{Figure_1},
	the per-node capacity is limited
	because of multiple hop transmissions \cite{Capacity_UAV_Relaying} in the aerial tier of UAV mesh networks.
Fortunately,
there is spatial separation between UAV mesh networks and ground networks \cite{cog_survey}.
Besides, in disaster management,
most of ground BSs are destroyed and the density of ground BS is small,
which creates the opportunity for UAVs to share
the spectrum of ground networks.
In such way, the capacity of UAV mesh networks can be improved.
Actually, the study of the spectrum sharing for UAV networks is initiated.
Zhang \emph{et al.} \cite{cog_zhangwei, book} initially
allowed drone
small cell networks to share the spectrum of cellular networks.
Lyu \emph{et al.} \cite{cog_zengyong}
proposed an orthogonal spectrum sharing scheme for UAV
networks.
The spectrum sharing
between UAV and radar was studied in \cite{cog_3d}.
The power control in
cognitive UAV networks was studied in \cite{cog_ee} to optimize energy efficiency.
Huang \emph{et al.} \cite{cog_routing} studied
the protocol design in
cognitive aerial networks.
Wang \emph{et al.} \cite{ss_JSAC}
addressed the trajectory planning and spectrum sharing scheme
for cognitive UAV networks.

Although the
spectrum sharing of UAV networks was studied,
very few of literatures studied
spectrum sharing between UAV mesh networks and
ground networks.
In this paper, considering two-dimensional (2D) and three-dimensional (3D) deployment of UAVs, we study the spectrum sharing between the
	A2A communications of UAVs
	and ground networks to enhance the
	capacity of A2A communications. The coverage performance of UAV mesh network is analyzed applying the stochastic geometry.
Directional antennas are employed to
improve coverage performance of UAV mesh network.
The performance improvement using directional antennas
is compared with UAV network
using omnidirectional antennas.
Finally, with the object function of maximizing
transmission capacity, the optimal altitude of UAVs is obtained.
It is noted that part of this work was published in
WCSP 2018 \cite{wcsp_hangzhou}.
The novel contributions include
study of spectrum sharing with 2D UAV network,
directional A2A and air-to-ground (A2G) transmission,
and the transition from 3D UAV network to 2D UAV network.
Specifically, we discover that derived results
of 3D UAV network transit to that of 2D UAV network
when the vertical range of 3D UAV deployment tends to 0.
Finally, more comprehensive simulation results are provided in this paper
compared with the conference paper.
Hence, this paper provides a complete study for
spectrum sharing of UAV mesh networks.

The rest of this paper is organized as follows.
In Section II, the system model of this paper is introduced.
The spectrum sharing of UAV
networks with omnidirectional transmission is analyzed in Section III.
The coverage probabilities of
UAV and ground networks with directional transmission are derived in Section IV.
The numerical results and analysis are
	provided in Section V.
Finally, Section VI provides the conclusion.
The key parameters and notations are listed in
Table \ref{tab_1}.

\section{System Model}

\subsection{Network model}

\begin{figure}
	\centering
	\includegraphics[width=0.48\textwidth]{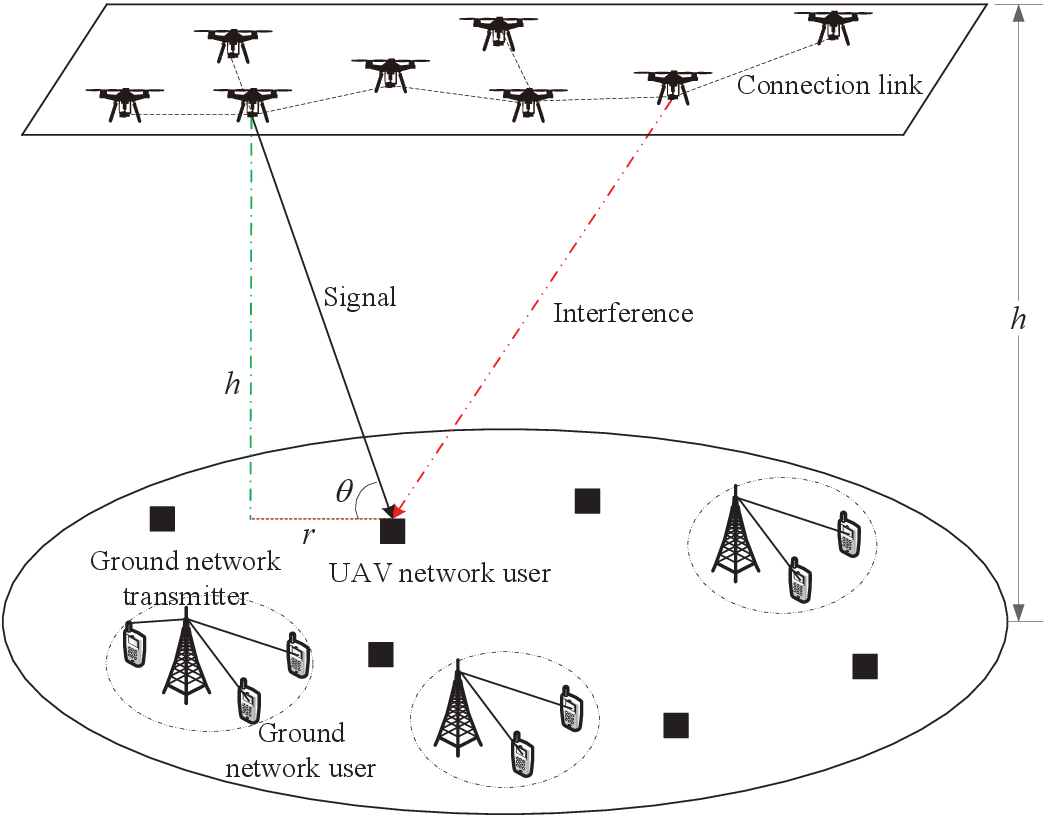}
	\caption{2D deployment of UAVs.}
	\label{Figure_2}
\end{figure}

\begin{figure}
	\centering
	\includegraphics[width=0.48\textwidth]{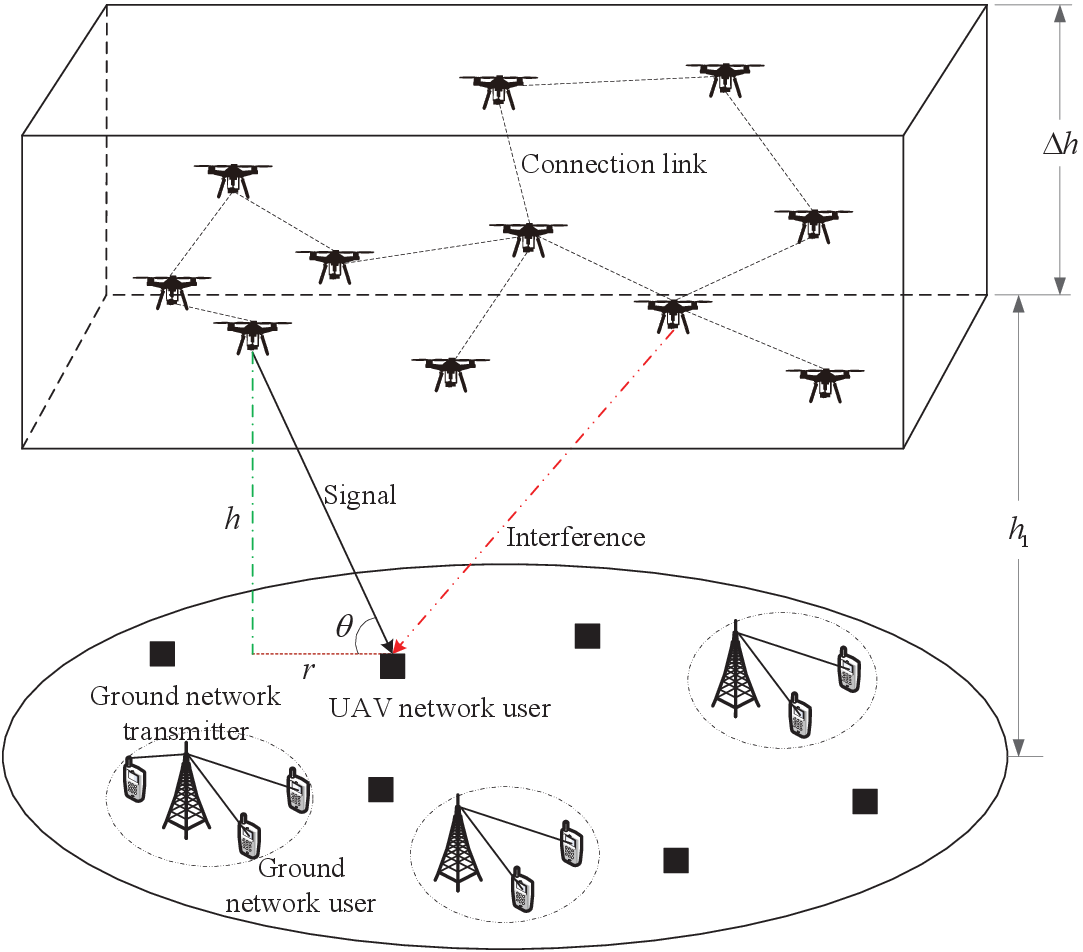}
	\caption{3D deployment of UAVs.}
	\label{Figure_3}
\end{figure}

As illustrated in Fig. \ref{Figure_2}
and Fig. \ref{Figure_3},
2D and 3D deployment of UAVs are considered.
For 2D scenario,
the distributions of the
UAVs and transmitters (TXs) of ground network
follow 2D homogeneous PPP
with densities ${\lambda_{2Du}}$
and ${\lambda _d}$, respectively.
The UAVs are deployed above the ground with height $h$.
In 3D scenario,
the distribution of
TXs of ground network is the same
to the scenario in Fig. \ref{Figure_2}.
The UAVs are distributed following a 3D
homogeneous PPP with density
${\lambda_{3Du}}$.
UAVs are deployed in the space between two planes with heights
$h_1$ and $h_1 + \Delta h$, respectively.

\subsection{Channel model}

Path-loss exponents of ground-to-ground (G2G) and
A2G links are
$\alpha_d$ and $\alpha_u$, respectively.
The power gain of small scale fading is modeled
using an exponentially distributed random variable with unit mean.
A Gaussian distribution random variable with mean
zero and variance $N$ is used to
model the power of additive white Gaussian noise.
UAV and TX of ground network transmit with power $P_u$ and $P_d$, respectively.
The received power of UAV network user in the cases of
line-of-sight (LoS) and non-line-of-sight (NLoS) transmissions
is \cite{ss_D2D, UAV_Channel}
\begin{equation}\label{Eq:1}
{P_{r,g}} = \left\{ {\begin{array}{*{20}{c}}
{{P_u}{{\left| {{d_{ug}}} \right|}^{ - {\alpha _u}}}}&{{\rm{LoS}}}\\
{\eta {P_u}{{\left| {{d_{ug}}} \right|}^{ - {\alpha _u}}}}&{{\rm{NLoS}}}
\end{array}} \right.,
\end{equation}
where $\eta$
is a diminution
factor \cite{UAV_Channel} and ${{d_{ug}}}$ is the distance between
UAV and the associated user.
The A2G link is LoS with probability \cite{UAV_Channel}.
\begin{equation}
{P_{\rm{LoS}}} = \frac{1}{{1 + C\exp \left( { - B(\theta  - C)} \right)}},
\end{equation}
where $\theta$ is the elevation angle,
$B$ and $C$ are environment dependent constants.

\subsection{Directional antenna model}

Directional antennas are
applied in A2A and
A2G links
to reduce interference.
The antenna gain expressed in the unit of dB is \cite{Directional_Antenna_Model, liuyunfeng}

\begin{equation}\label{gain}
\begin{aligned}
&G(\alpha ,{\theta _{ - 3dB}}) = \left\{ {\begin{array}{*{20}{c}}
	{{G_0} - 3.01 \cdot {{(\frac{{2\alpha }}{{{\theta _{ - 3dB}}}})}^2},}&{{0^\circ } \le \alpha  \le \frac{{{\theta _{ml}}}}{2}},\\
	{{G_{sl}},}&{\frac{{{\theta _{ml}}}}{2} \le \alpha  \le {{180}^\circ }},
	\end{array}} \right.\\
&{G_0} = 10\log ({(\frac{{1.6162}}{{\sin ({\theta _{ - 3dB}}/2)}})^2}),\\
 &{\theta _{ml}} = 2.6 \cdot {\theta _{ - 3dB}},\\
&{G_{sl}} =  - 0.4111 \cdot \ln ({\theta _{ - 3dB}}) - 10.597,
\end{aligned}
\end{equation}
where $G_0$ is the maximum antenna gain,
$G_{sl}$ is the antenna gain of side lobe,
$\theta _{ - 3dB}$ is the half power bandwidth
of directional antenna,
$\alpha$ is the angle between the line connecting TX and receiver (RX) and
the center line of transmit beam,
and $\theta _{ml}$ is the angle of main lobe.

\section{Spectrum Sharing with Omnidirectional Transmission}

In order to study the characteristics of spectrum sharing between UAV mesh networks and ground networks, we introduce the concept of the coverage probability. The coverage probability is defined as the probability of
successful communication for ground network user and UAV network user. In this paper, considering that UAV network users and ground network users are the same type of users, the threshold of signal-to-interference-plus-noise ratio (SINR) for successful communication is the same, denoted by $\beta$. $P_1$ denotes the coverage probability of typical ground network user. Its main interference includes ground network transmitters and UAV networks under spectrum sharing. $P_2$ denotes the coverage probability of typical UAV network user. The main interference comes from the UAV network air-to-ground communication links.

In this section, with 2D and 3D deployment of UAVs,
the performance of spectrum sharing
is studied.
When $\Delta h \to 0$, 3D deployment of UAVs reduces to
2D deployment of UAVs.
In this situation, we reveal that
coverage probability with 3D deployment of UAVs turns into
coverage probability with 2D deployment of UAVs.

\subsection{Spectrum sharing of 2D UAV network}
\newcounter{mytempeqncnt}
\begin{figure*}[ht]
\normalsize
\begin{equation}\label{eq_2DG}
\begin{aligned}
{L_{I_{gu}^c}}( {\frac{{\beta {d_0}^{{\alpha _d}}}}{{{P_d}}}} ) = \exp ( { - \frac{{2\lambda_d {\pi ^2}{{( \beta  )}^{2/{\alpha _d}}}{d_0}^2}}{{{\alpha _d}\sin ( {2\pi /{\alpha _d}} )}}} ).
\end{aligned}
\end{equation}
\begin{equation}\label{eq_2DG_LOS}
\begin{aligned}
{L_{{I_{u,\rm{LoS}}}}}( {\frac{{\beta d_0^{{\alpha _d}}}}{{{P_d}}}} )& = \exp ( { - {\lambda _{2Du}}\int\limits_V {( {1 - \frac{1}{{1 + \frac{{\beta d_0^{{\alpha _d}}}}{{{P_d}}}{P_u}{P_{\rm{LoS}}}x_i^{ - {\alpha _u}}}}} )} dx} )\\
&= \exp ( { - 2\pi {\lambda _{2Du}}\int_0^\infty  {( {1 - \frac{1}{{1 + \frac{{\beta d_0^{{\alpha _d}}}}{{{P_d}}}{P_u}{{(\sqrt {{r^2} + {h^2}} )}^{ - {\alpha _u}}}\frac{1}{{1 + C\exp ( { - B(\frac{{180}}{\pi }arctan(h/r) - C)} )}}}}} )rdr} } )\\
&= \exp ( { - 2\pi {\lambda _{2Du}}{H_1}( {\beta ,{d_0},h,{\alpha _d},{\alpha _u}} )} ),
\end{aligned}
\end{equation}

\begin{equation}\label{eq_2DG_NLOS}
\begin{aligned}
{L_{{I_{u,\rm{NLoS}}}}}( {\frac{{\beta d_0^{{\alpha _d}}}}{{{P_d}}}} ) &= \exp ( { - 2\pi {\lambda _{2Du}}\int_0^\infty  {( {1 - \frac{1}{{1 + \frac{{\beta d_0^{{\alpha _d}}}}{{{P_d}}}{P_u}{{(\sqrt {{r^2} + {h^2}} )}^{ - {\alpha _u}}}\eta{(1{-}\frac{1}{{1 + C\exp ( { - B(\frac{{180}}{\pi }arctan(h/r) - C)} )}})}}}} ){r}dr} } )\\
&= \exp ( { - 2\pi {\lambda _{2Du}}{H_2}( {\beta ,{d_0},h,{\alpha _d},{\alpha _u}} )} ),
\end{aligned}
\end{equation}
where

\begin{equation}\label{eq_h1}
{H_1}( {\beta ,{d_0},h,{\alpha _d},{\alpha _u}} ) = \int_0^\infty  {( {1 - \frac{1}{{1 + \frac{{\beta d_0^{{\alpha _d}}}}{{{P_d}}}{P_u}{{(\sqrt {{r^2} + {h^2}} )}^{ - {\alpha _u}}}\frac{1}{{1 + C\exp ( { - B(\frac{{180}}{\pi }arctan(h/r) - C)} )}}}}} ){r}dr},
\end{equation}

\begin{equation}\label{eq_h2}
{H_2}( {\beta ,{d_0},h,{\alpha _d},{\alpha _u}} ) = \int_0^\infty  {( {1 - \frac{1}{{1 + \frac{{\beta d_0^{{\alpha _d}}}}{{{P_d}}}{P_u}{{(\sqrt {{r^2} + {h^2}} )}^{ - {\alpha _u}}}\eta{(1{-}\frac{1}{{1 + C\exp ( { - B(\frac{{180}}{\pi }arctan(h/r) - C)} )}})}}}} ){r}dr}.
\end{equation}
\hrulefill
\end{figure*}

\subsubsection{Performance of ground network}

For ground network,
the signal to interference plus noise ratio (SINR) of
the typical user, namely, the user at the origin $\{ \bf{0}\}$
is
\begin{equation}
{\gamma_{gu}}=\frac{{{P_d}d_0^{ - {\alpha _d}}{g_0}}}{{I_{gu}^c + {I_u} + N}},
\end{equation}
where $d_0$ denotes the distance between
the user at the origin $\{ \bf{0}\}$
and the associated BS,
${I_{gu}^c}$ is the aggregated interference
received by the typical user from the BSs of ground network,
${{I_u}}$ is the aggregated interference received by the typical user from UAVs,
and $g_0$ is the power gain of small scale channel fading between
	the user at the origin $\{ \bf{0}\}$
	and the associated BS.
\begin{equation}\label{I_gu}
I_{gu}^c = \sum\limits_{{d_i} \in {\Phi _d}\backslash \{ \bf{0}\} } {{P_d}} d_i^{ - {\alpha _d}}{g_i},
\end{equation}
\begin{equation}
\begin{aligned}
{I_u} & = {I_{u,LoS}} + {I_{u,{\rm{NLoS}}}}\\
& = \sum\limits_{{x_i} \in {\Phi _u}} {{P_{{\rm{LoS}}}}{P_u}} x_i^{ - {\alpha _d}}{g_i} + \sum\limits_{{x_i} \in {\Phi _u}} {\left( {1 - {P_{{\rm{LoS}}}}} \right)\eta {P_u}} x_i^{ - {\alpha _d}}{g_i},
\end{aligned}
\end{equation}
where
$g_i$ is the power gain of small scale channel fading of $i$th interference link,
$x_i$ is the distance between $i$th UAV and the user at the origin $\{ \bf{0}\}$,
and $d_i$ is the distance between $i$th BS of ground network and
the user at the origin $\{ \bf{0}\}$.
For the ground network, the coverage probability of typical user is
\begin{equation}\label{p1}
\begin{aligned}
{P_1} &= P( {\gamma_{gu}} > \beta),\\
& = \exp ( { - \frac{{\beta d_0^{{\alpha _d}}( {I_{gu}^c + {I_u} + N} )}}{{{P_d}}}} )\\
& = \exp ( { - \frac{{\beta d_0^{{\alpha _d}}I_{gu}^c}}{{{P_d}}}} )\exp ( { - \frac{{\beta d_0^{{\alpha _d}}{I_u}}}{{{P_d}}}} )\exp ( { - \frac{{\beta d_0^{{\alpha _d}}N}}{{{P_d}}}} )\\
& = {L_{I_{gu}^c}}( {\frac{{\beta d_0^{{\alpha _d}}}}{{{P_d}}}} ){L_{{I_u}}}( {\frac{{\beta d_0^{{\alpha _d}}}}{{{P_d}}}} )\exp ( { - \frac{{\beta d_0^{{\alpha _d}}N}}{{{P_d}}}} ),
\end{aligned}
\end{equation}
where ${L_A}(*)$ represents Laplace transform of $A$.
With the channel model in (\ref{Eq:1}),
we have $I_u={I_{u,{\rm{LoS}}}}+{I_{u,{\rm{NLoS}}}}$.
The Laplace transform of
$I_u$ is
\begin{equation}
 {L_{{I_u}}}( {\frac{{\beta d_0^{{\alpha _d}}}}{{{P_d}}}} ) = {L_{{I_{u,{\rm{LoS}}}}}}( {\frac{{\beta d_0^{{\alpha _d}}}}{{{P_d}}}} ){L_{{I_{u,N{\rm{LoS}}}}}}( {\frac{{\beta d_0^{{\alpha _d}}}}{{{P_d}}}} ).
\end{equation}

Since $g_i$ and PPP
are independent,
we have
\begin{equation}\label{ldlos}
\begin{aligned}
& {L_{{I_{u,{\rm{LoS}}}}}}( {\frac{{\beta d_0^{{\alpha _d}}}}{{{P_d}}}} ) = {E_{{I_{u,{\rm{LoS}}}}}}[ {\exp ( {\frac{{\beta d_0^{{\alpha _d}}}}{{{P_d}}}{I_{u,{\rm{LoS}}}}} )} ]\\
& = {E_{{g_i},{\Phi _u}}}[ {\prod\limits_{{x_i} \in {\Phi _u}\backslash \{ \bf{0}\} } {\exp ( {\frac{{\beta d_0^{{\alpha _d}}}}{{{P_d}}}{g_i}{P_u}{P_{{\rm{LoS}}}}x_i^{ - {\alpha _u}}} )} } ]\\
& = {E_{{\Phi _u}}}[ {\prod\limits_{{x_i} \in {\Phi _u}\backslash \{ \bf{0}\} } {{E_{{g_i}}}[\exp ( {\frac{{\beta d_0^{{\alpha _d}}}}{{{P_d}}}{P_u}{P_{{\rm{LoS}}}}x_i^{ - {\alpha _u}}} )]} } ]\\
& = {E_{{\Phi _u}}}[ {\prod\limits_{{x_i} \in {\Phi _u}\backslash \{ \bf{0}\} } {\frac{1}{{1 + \frac{{\beta d_0^{{\alpha _d}}}}{{{P_d}}}{P_u}{P_{{\rm{LoS}}}}x_i^{ - {\alpha _u}}}}} } ],
\end{aligned}
\end{equation}
and
\begin{equation}\label{ldnlos}
\begin{aligned}
& {L_{{I_{u,{\rm{NLoS}}}}}}( {\frac{{\beta d_0^{{\alpha _d}}}}{{{P_d}}}} )\\
& = {E_{{\Phi _u}}}[ {\prod\limits_{{x_i} \in {\Phi _u}\backslash \{ \bf{0}\} } {\frac{1}{{1 + \frac{{\beta d_0^{{\alpha _d}}}}{{{P_d}}}{P_u}(1 - {P_{{\rm{LoS}}}})\eta x_i^{ - {\alpha _u}}}}} } ].
\end{aligned}
\end{equation}

Hence, we have derived ${L_{{I_u}}}( {\frac{{\beta d_0^{{\alpha _d}}}}{{{P_d}}}} )$.
\begin{equation}
\begin{aligned}
& {L_{{I_u}}}( {\frac{{\beta d_0^{{\alpha _d}}}}{{{P_d}}}} ) = {E_{{\Phi _u}}}[ {\prod\limits_{{x_i} \in {\Phi _u}} {\frac{1}{{1 + \frac{{\beta d_0^{{\alpha _d}}}}{{{P_d}}}{P_u}{P_{{\rm{LoS}}}}x_i^{ - {\alpha _u}}}}} } ] \\
&\times {E_{{\Phi _u}}}[ {\prod\limits_{{x_i} \in {\Phi _u}} {\frac{1}{{1 + \frac{{\beta d_0^{{\alpha _d}}}}{{{P_d}}}{P_u}(1 - {P_{{\rm{LoS}}}})\eta x_i^{ - {\alpha _u}}}}} } ].
\end{aligned}
\end{equation}

The probability generating function of PPP is as follows \cite{cog_zhangwei}.
\begin{equation}\label{eq_pgf}
E( {\prod\limits_{{x_i} \in \Phi } {f( x )} } ) = \exp ( { - {\lambda _d}\int\limits_V {[ {1 - f( x )} ]} dx} ),
\end{equation}

Hence, we have
${L_{I_{gu}^c}}( {\frac{{\beta d_0^{{\alpha _d}}}}{{{P_d}}}} )$
as in (\ref{eq_2DG}) \cite{123}.
${L_{{I_u}}}( {\frac{{\beta d_0^{{\alpha _d}}}}{{{P_d}}}} )$
is derived using (\ref{eq_2DG_LOS}) and (\ref{eq_2DG_NLOS}),
where ${H_1}( {\beta ,{d_0},h,{\alpha _d},{\alpha _u}} )$ and ${H_2}( {\beta ,{d_0},h,{\alpha _d},{\alpha _u}} )$
are provided in (\ref{eq_h1}) and (\ref{eq_h2}), respectively.

\begin{figure*}[ht]
\normalsize
\begin{equation}\label{eq_2uiasduia}
\begin{aligned}
{L_{I_{u,\rm{LoS}}^c}}( {\frac{{\beta x_0^{{\alpha _u}}}}{{{P_u}}}} ) &= \exp ( { - {\lambda _{2Du}}\int\limits_V {( {1 - \frac{1}{{1 + \beta x_0^{{\alpha _u}}{P_{\rm{LoS}}}x_i^{ - {\alpha _u}}}}} )} dx} )\\
&= \exp ( { - 2\pi {\lambda _{2Du}}\int_0^\infty  {( {1 - \frac{1}{{1 + \beta x_0^{{\alpha _u}}{{(\sqrt {{r^2} + {h^2}} )}^{ - {\alpha _u}}}\frac{1}{{1 + C\exp ( { - B(\frac{{180}}{\pi }arctan(h/r) - C)} )}}}}} )rdr} } )\\
&= \exp ( { - 2\pi {\lambda _{2Du}}{H_3}( {\beta ,{x_0},h,{\alpha _u}} )} ),
\end{aligned}
\end{equation}

\begin{equation}\label{eq_2ugsg}
\begin{aligned}
{L_{I_{u,{\rm{NLoS}}}^c}}( {\frac{{\beta x_0^{{\alpha _u}}}}{{{P_u}}}} ) &= \exp ( { - 2\pi {\lambda _{2Du}}\int_0^\infty  {( {1 - \frac{1}{{1 + \beta x_0^{{\alpha _u}}{{(\sqrt {{r^2} + {h^2}} )}^{ - {\alpha _u}}}\eta{(1{-}\frac{1}{{1 + C\exp ( { - B(\frac{{180}}{\pi }arctan(h/r) - C)} )}}})}}} )rdr} } )\\
&= \exp ( { - 2\pi {\lambda _{2Du}}{H_4}( {\beta ,{x_0},h,{\alpha _u}} )} ),
\end{aligned}
\end{equation}
where
\begin{equation}\label{eq_h3}
{H_3}( {\beta ,{x_0},h,{\alpha _u}} ) = \int_0^\infty  {( {1 - \frac{1}{{1 + \beta x_0^{{\alpha _u}}{{(\sqrt {{r^2} + {h^2}} )}^{ - {\alpha _u}}}\frac{1}{{1 + C\exp ( { - B(\frac{{180}}{\pi }arctan(h/r) - C)} )}}}}} )rdr},
\end{equation}

\begin{equation}\label{eq_h4}
{H_4}( {\beta ,{x_0},h,{\alpha _u}} ){=}\int_0^\infty  {( {1 - \frac{1}{{1 + \beta x_0^{{\alpha _u}}{{(\sqrt {{r^2} + {h^2}} )}^{ - {\alpha _u}}}\eta{(1{-}\frac{1}{{1 + C\exp ( { - B(\frac{{180}}{\pi }arctan(h/r) - C)} )}}})}}} )rdr}.
\end{equation}
\hrulefill \vspace*{4pt}
\end{figure*}

\subsubsection{Performance of UAV network}

For UAV network, the coverage probability of its typical user is
\begin{equation}
{P_2} = P( {\gamma_{uu} > \beta } ),
\end{equation}
where $\beta$
is SINR threshold and $\gamma_{uu}$ is the
SINR of the typical user in UAV network.
Applying the channel model in (\ref{Eq:1}),
${P_2}$ is expressed as
\begin{equation}\label{eq_coverage_probability_UAV_user}
\begin{aligned}
P_2 & = {P_{{\text{LoS}}}}P( {\frac{{{P_u}x_0^{ - {\alpha _u}}{g_i}}}{{I_u^c}} > \beta } ) + {P_{{\text{NLoS}}}}P( {\frac{{\eta {P_u}x_0^{ - {\alpha _u}}{g_i}}}{{I_u^c}} > \beta } )\\
& = {P_{{\text{LoS}}}}\exp ( { - \frac{{\beta x_0^{{\alpha _u}}I_u^c}}{{{P_u}}}} ) + (1 - {P_{{\text{LoS}}}})\exp ( { - \frac{{\beta x_0^{{\alpha _u}}I_u^c}}{{\eta {P_u}}}} )\\
& = {P_{{\text{LoS}}}}{L_{I_u^c}}( { - \frac{{\beta x_0^{{\alpha _u}}}}{{{P_u}}}} ) + (1 - {P_{{\text{LoS}}}}){L_{I_u^c}}( { - \frac{{\beta x_0^{{\alpha _u}}}}{{\eta {P_u}}}} ),
\end{aligned}
\end{equation}
where $x_0$ is the distance between the associated UAV BS and the typical user.
$I_u^c$ is the aggregated interference received by the typical user from UAVs,
which is expressed as follows.
\begin{equation}\label{eq_yuasd}
\begin{aligned}
I_u^c & = I_{_{u,{\text{LoS}}}}^c + I_{_{u,{\text{NLoS}}}}^c\\
&  = \sum\limits_{{x_i} \in {\Phi _u}\backslash \{ \bf{0}\} } {{P_{{\text{LoS}}}}{P_u}} x_i^{ - {\alpha _u}}{g_i} + \\
& \sum\limits_{{x_i} \in {\Phi _u}\backslash \{ \bf{0}\} } {( {1 - {P_{{\text{LoS}}}}} )\eta {P_u}} x_i^{ - {\alpha _u}}{g_i}.
\end{aligned}
\end{equation}

The Laplace transform of $I_u^c$ is
\begin{equation}\label{eq_yuiasyuda}
\begin{aligned}
{L_{I_u^c}}( {\frac{{\beta x_0^{{\alpha _u}}}}{{{P_u}}}} ) & = {L_{I_{u,{\text{LoS}}}^c}}( {\frac{{\beta x_0^{{\alpha _u}}}}{{{P_u}}}} ){L_{I_{u,{\text{NLoS}}}^c}}( {\frac{{\beta x_0^{{\alpha _u}}}}{{{P_u}}}} )\\
& = {E_{{\Phi _u}}}[ {\prod\limits_{{x_i} \in {\Phi _u}\backslash \{ \bf{0}\} } {\frac{1}{{1 + \beta x_0^{{\alpha _u}}{P_{{\text{LoS}}}}x_i^{ - {\alpha _u}}}}} } ] \times  \hfill \\
& {E_{{\Phi _u}}}[ {\prod\limits_{{x_i} \in {\Phi _u}\backslash \{ \bf{0}\} } {\frac{1}{{1 + \beta x_0^{{\alpha _u}}(1 - {P_{{\text{LoS}}}})\eta x_i^{ - {\alpha _u}}}}} } ].
\end{aligned}
\end{equation}

${L_{I_{u,{\text{LoS}}}^c}}( {\frac{{\beta x_0^{{\alpha _u}}}}{{{P_u}}}} )$
and ${L_{I_{u,{\text{NLoS}}}^c}}( {\frac{{\beta x_0^{{\alpha _u}}}}{{{P_u}}}} )$ are derived
in (\ref{eq_2uiasduia}) and (\ref{eq_2ugsg}), where
${H_3}( {\beta ,{x_0},h,{\alpha _u}} )$ and ${H_4}( {\beta ,{x_0},h,{\alpha _u}} )$ are provided in (\ref{eq_h3}) and (\ref{eq_h4}), respectively.

\subsection{Spectrum sharing of 3D UAV network}

With UAVs deployed in 3D space,
height dimension needs to be considered.
For the ground network, the coverage probability of typical user, denoted by
${P_1}$, is as follows according to (\ref{p1}).
\begin{equation}\label{p11}
\begin{aligned}
{P_1}  = {L_{I_{gu}^c}}( {\frac{{\beta d_0^{{\alpha _d}}}}{{{P_d}}}} ){L_{{I_u}}}( {\frac{{\beta d_0^{{\alpha _d}}}}{{{P_d}}}} )\exp ( { - \frac{{\beta d_0^{{\alpha _d}}N}}{{{P_d}}}} ).
\end{aligned}
\end{equation}

${L_{{I_u}}}( {\frac{{\beta d_0^{{\alpha _d}}}}{{{P_d}}}} )$ is derived in (\ref{eq_3DG_LOS}) and (\ref{eq_3DG_NLOS}),
where ${H_5}( {\beta ,{d_0},h,{\alpha _d},{\alpha _u}} )$ and ${H_6}( {\beta ,{d_0},h,{\alpha _d},{\alpha _u}} )$
are derived in (\ref{eq_h5}) and (\ref{eq_h6}).

For UAV network, the coverage probability of typical user is denoted by ${P_2}$.
According to (\ref{eq_coverage_probability_UAV_user}) and (\ref{eq_yuiasyuda}),
we have
\begin{equation}
\begin{aligned}
{P_2} = {P_{\rm{LoS}}}{L_{I_{u,\rm{LoS}}^c}}\left( {\frac{{\beta x_0^{{\alpha _u}}}}{{{P_u}}}} \right){L_{I_{u,\rm{NLoS}}^c}}\left( {\frac{{\beta x_0^{{\alpha _u}}}}{{{P_u}}}} \right) + \\
(1 - {P_{\rm{LoS}}}){L_{I_{u,\rm{LoS}}^c}}\left( {\frac{{\beta x_0^{{\alpha _u}}}}{{{P_u}}}} \right){L_{I_{u,\rm{NLoS}}^c}}\left( {\frac{{\beta x_0^{{\alpha _u}}}}{{{P_u}}}} \right).
\end{aligned}
\end{equation}

${L_{I_{u,{\text{LoS}}}^c}}( {\frac{{\beta x_0^{{\alpha _u}}}}{{{P_u}}}} )$
and ${L_{I_{u,{\text{NLoS}}}^c}}( {\frac{{\beta x_0^{{\alpha _u}}}}{{{P_u}}}} )$
are derived
in (\ref{eq_3uiasduia}) and (\ref{eq_3ugsg}), where
${H_7}( {\beta ,{x_0},h,{\alpha _u}} )$ and ${H_8}( {\beta ,{x_0},h,{\alpha _u}} )$ are provided in (\ref{eq_h7}) and (\ref{eq_h8}), respectively.

\begin{figure*}[ht]
	\normalsize
	\begin{equation}\label{eq_3DG_LOS}
	\begin{aligned}
	{L_{{I_{u,\rm{LoS}}}}}( {\frac{{\beta d_0^{{\alpha _d}}}}{{{P_d}}}} ) & = \exp ( { - {\lambda _{3Du}}\int\limits_V {( {1 - \frac{1}{{1 + \frac{{\beta d_0^{{\alpha _d}}}}{{{P_d}}}{P_u}{P_{\rm{LoS}}}x_i^{ - {\alpha _u}}}}} )} dx} )\\
	& = \exp ( { - 2\pi {\lambda _{3Du}}\int_{h1}^{h2} {\int_0^\infty  {( {1 - \frac{1}{{1 + \frac{{\beta d_0^{{\alpha _d}}}}{{{P_d}}}{P_u}{{(\sqrt {{r^2} + {z^2}} )}^{ - {\alpha _u}}}\frac{1}{{1 + C\exp ( { - B(\frac{{180}}{\pi }\arctan (z/r) - C)} )}}}}} )} } rdrdz} )\\
	& = \exp ( { - 2\pi {\lambda _{3Du}}{H_5}( {\beta ,{d_0},h,{\alpha _d},{\alpha _u}} )} ),
	\end{aligned}
	\end{equation}
	
	\begin{equation}\label{eq_3DG_NLOS}
	\begin{aligned}
	{L_{{I_{u,{\text{NLoS}}}}}}( {\frac{{\beta d_0^{{\alpha _d}}}}{{{P_d}}}} ) & = \exp ( { - 2\pi {\lambda _{3Du}}\int_{h1}^{h2} {\int_0^\infty  {( {1 - \frac{1}{{1 + \frac{{\beta d_0^{{\alpha _d}}}}{{{P_d}}}{P_u}\eta {{(\sqrt {{r^2} + {z^2}} )}^{ - {\alpha _u}}}(1 - \frac{1}{{1 + C\exp ( { - B(\frac{{180}}{\pi }\arctan (z/r) - C)} )}})}}} )} } rdrdz} )\\
	& = \exp ( { - 2\pi {\lambda _{3Du}}{H_6}( {\beta ,{d_0},h,{\alpha _d},{\alpha _u}} )} ),
	\end{aligned}
	\end{equation}
	where
	\begin{equation}\label{eq_h5}
	{H_5}( {\beta ,{d_0},h,{\alpha _d},{\alpha _u}} ) = \int_{h1}^{h2} {\int_0^\infty  {( {1 - \frac{1}{{1 + \frac{{\beta d_0^{{\alpha _d}}}}{{{P_d}}}{P_u}{{(\sqrt {{r^2} + {z^2}} )}^{ - {\alpha _u}}}\frac{1}{{1 + C\exp ( { - B(\frac{{180}}{\pi }\arctan (z/r) - C)} )}}}}} )} } rdrdz,
	\end{equation}
	
	\begin{equation}\label{eq_h6}
	{H_6}( {\beta ,{d_0},h,{\alpha _d},{\alpha _u}} ) = \int_{h1}^{h2} {\int_0^\infty  {( {1 - \frac{1}{{1 + \frac{{\beta d_0^{{\alpha _d}}}}{{{P_d}}}{P_u}\eta {{(\sqrt {{r^2} + {z^2}} )}^{ - {\alpha _u}}}(1 - \frac{1}{{1 + C\exp ( { - B(\frac{{180}}{\pi }\arctan (z/r) - C)} )}})}}} )} } rdrdz.
	\end{equation}

\begin{equation}\label{eq_3uiasduia}
\begin{aligned}
{L_{I_{u,{\rm{LoS}}}^c}}( {\frac{{\beta x_0^{{\alpha _u}}}}{{{P_u}}}} ) & = \exp ( { - {\lambda _{3Du}}\int\limits_V {( {1 - \frac{1}{{1 + \beta x_0^{{\alpha _u}}{P_{{\rm{LoS}}}}x_i^{ - {\alpha _u}}}}} )} dx} )\\
& = \exp ( { - {\lambda _{3Du}}\int_{h1}^{h2} {\int_0^{2\pi } {\int_0^\infty  {( {1 - \frac{1}{{1 + \beta x_0^{{\alpha _u}}{{(\sqrt {{r^2} + {z^2}} )}^{ - {\alpha _u}}}\frac{1}{{1 + C\exp ( { - B(\frac{{180}}{\pi }\arctan (z/r) - C)} )}}}}} )} } } rdrd\phi dz} )\\
& = \exp ( { - 2\pi {\lambda _{3Du}}{H_7}( {\beta ,{x_0},h,{\alpha _u}} )} ),
\end{aligned}
\end{equation}

\begin{equation}\label{eq_3ugsg}
\begin{aligned}
{L_{I_{u,{\rm{NLoS}}}^c}}( {\frac{{\beta x_0^{{\alpha _u}}}}{{{P_u}}}} ) & = \exp ( { - 2\pi {\lambda _{3Du}}\int_{h1}^{h2} {\int_0^\infty  {( {1 - \frac{1}{{1 + \beta x_0^{{\alpha _u}}\eta {{(\sqrt {{r^2} + {z^2}} )}^{ - {\alpha _u}}}(1 - \frac{1}{{1 + C\exp ( { - B(\frac{{180}}{\pi }arctan(z/r) - C)} )}})}}} )} } rdrdz} )\\
& = \exp ( { - 2\pi {\lambda _{3Du}}{H_8}( {\beta ,{x_0},h,{\alpha _u}} )} ),
\end{aligned}
\end{equation}
where
\begin{equation}\label{eq_h7}
{H_7}( {\beta ,{x_0},h,{\alpha _u}} ) = \int_{h1}^{h2} {\int_0^\infty  {( {1 - \frac{1}{{1 + \beta x_0^{{\alpha _u}}{{(\sqrt {{r^2} + {z^2}} )}^{ - {\alpha _u}}}\frac{1}{{1 + C\exp ( { - B(\frac{{180}}{\pi }\arctan (z/r) - C)} )}}}}} )} } rdrdz,
\end{equation}

\begin{equation}\label{eq_h8}
{H_8}( {\beta ,{x_0},h,{\alpha _u}} ) = \int_{h1}^{h2} {\int_0^\infty  {( {1 - \frac{1}{{1 + \beta x_0^{{\alpha _u}}\eta {{(\sqrt {{r^2} + {z^2}} )}^{ - {\alpha _u}}}(1 - \frac{1}{{1 + C\exp ( { - B(\frac{{180}}{\pi }\arctan (z/r) - C)} )}})}}} )} } rdrdz.
\end{equation}
 \hrulefill
\end{figure*}

\subsection{Transition from 3D UAV network to 2D UAV network}

In 3D UAV network,
UAVs are deployed in the space with
vertical range $\Delta h$.
When $\Delta h \to 0$,
3D UAV network becomes 2D UAV network.
In this situation,
UAVs are deployed on a plane with height ${h_1}$
and area ${S}$.
Supposing that the total number of UAVs is $N$,
we have
\begin{equation}
L_{I_{u,\rm{LoS}}^c}^{2D}\left( {\frac{{\beta x_0^{{\alpha _u}}}}{{{P_u}}}} \right){=}\exp ( - 2\pi \frac{N}{S}f\left( {{h_1}} \right)),
\end{equation}
and
\begin{equation}
L_{_{I_{u,\rm{LoS}}^c}}^{3D}\left( {\frac{{\beta x_0^{{\alpha _u}}}}{{{P_u}}}} \right){=
\rm{exp}}( - 2\pi \frac{N}{{S \cdot \Delta h}}\int_{h1}^{h1 + \Delta h} {f(z)} dz).
\end{equation}

According to L'H$\hat{\rm{o}}$pital's rule,
we have
\begin{equation}\label{eq_yausdyua1}
\begin{aligned}
L_{_{I_{u,\rm{LoS}}^c}}^{3D}\left( {\frac{{\beta x_0^{{\alpha _u}}}}{{{P_u}}}} \right)
&\mathop {=}\limits^{\Delta h \to 0} {\rm{exp}}( - 2\pi \frac{N}{{S \cdot \Delta h}}\int_{h1}^{h1 + \Delta h} {f(z)} dz)\\
&\mathop {=}\limits^{\Delta h \to 0} {\rm{exp}}( - 2\pi \frac{N}{S}f(h1))\\
&\mathop {=}\limits^{\Delta h \to 0} L_{I_{u,\rm{LoS}}^c}^{2D}\left( {\frac{{\beta x_0^{{\alpha _u}}}}{{{P_u}}}} \right).
\end{aligned}
\end{equation}

According to (\ref{eq_yausdyua1}),
when $\Delta h \to 0$,
the coverage probabilities with 3D UAV deployment tend to
the results
with 2D UAV deployment.

\section{Spectrum Sharing with Directional Transmission}

In this section,
directional transmission is
considered to improve the performance of
spectrum sharing.
Antenna gain is provided in (\ref{gain}).

\subsubsection{Spectrum sharing of 2D UAV network}

When UAV adopts directional transmission,
${I_{gu}^c}$ is (\ref{I_gu})
and ${P_1}$ is (\ref{p1}).
Then ${I_{u}}$ is derived as follows.
\begin{equation}
\begin{aligned}
{I_u} & = {I_{u,\rm{LoS}}} + {I_{u,\rm{NLoS}}}\\
& = \sum\limits_{{x_i} \in {\Phi _u}} {{P_{\rm{LoS}}}{P_u}} x_i^{ - {\alpha _d}}{g_i}G\left( {{\varphi _i}} \right) + \\
&\sum\limits_{{x_i} \in {\Phi _u}} {\left( {1 - {P_{\rm{LoS}}}} \right)\eta {P_u}} x_i^{ - {\alpha _d}}{g_i}G\left( {{\varphi _i}} \right).
\end{aligned}
\end{equation}

Then ${L_{{I_{u,\rm{LoS}}}}}\left( {\frac{{\beta d_0^{{\alpha _d}}}}{{{P_d}}}} \right)$ and ${L_{{I_{u,\rm{NLoS}}}}}\left( {\frac{{\beta d_0^{{\alpha _d}}}}{{{P_d}}}} \right)$ are modified as follows.
\begin{equation}
\begin{aligned}
{L_{{I_{u,\rm{LoS}}}}}\left( {\frac{{\beta d_0^{{\alpha _d}}}}{{{P_d}}}} \right) = {E_{{I_{u,\rm{LoS}}}}}\left[ {\exp \left( {\frac{{\beta d_0^{{\alpha _d}}}}{{{P_d}}}{I_{u,\rm{LoS}}}} \right)} \right]\\
= {E_{{\Phi _u}}}\left[ {\prod\limits_{{x_i} \in {\Phi _u}\backslash \{ 0\} } {\frac{1}{{1 + \frac{{\beta d_0^{{\alpha _d}}}}{{{P_d}}}{P_u}{P_{\rm{LoS}}}x_i^{ - {\alpha _u}}G\left( {{\varphi _i}} \right)}}} } \right].
\end{aligned}
\end{equation}

\begin{equation}
\begin{split}
&{L_{{I_{u,\rm{NLoS}}}}}\left( {\frac{{\beta d_0^{{\alpha _d}}}}{{{P_d}}}} \right)\\
& ={E_{{\Phi _u}}}\left[ {\prod\limits_{{x_i} \in {\Phi _u}\backslash \{ 0\} } {\frac{1}{{1 + \frac{{\beta d_0^{{\alpha _d}}}}{{{P_d}}}{P_u}(1 - {P_{\rm{LoS}}})\eta x_i^{ - {\alpha _u}}G\left( {{\varphi _i}} \right)}}} } \right].
\end{split}
\end{equation}

${L_{I_{u,{\text{LoS}}}}}( {\frac{{\beta d_0^{{\alpha _d}}}}{{{P_d}}}} )$
and ${L_{I_{u,{\text{NLoS}}}}}( {\frac{{\beta d_0^{{\alpha _d}}}}{{{P_d}}}} )$ can be derived using the probability generation function
in (\ref{eq_2gaind2dlos}) and (\ref{eq_2gaind2dnlos}),
where ${H_9}( {\beta ,{d_0},h,{\alpha _u}} )$ and ${H_{10}}( {\beta ,{d_0},h,{\alpha _u}} )$ are provided in (\ref{eq_h9}) and (\ref{eq_h10}), respectively.

For UAV network user, we have
\begin{equation}
\begin{aligned}
I_u^c & = I_{_{u,\rm{LoS}}}^c + I_{_{u,\rm{NLoS}}}^c\\
& = \sum\limits_{{x_i} \in {\Phi _u}\backslash \{ 0\} } {{P_{\rm{LoS}}}{P_u}} x_i^{ - {\alpha _d}}{g_i}G\left( {{\phi _i}} \right) + \\
&\sum\limits_{{x_i} \in {\Phi _u}\backslash \{ 0\} } {\left( {1 - {P_{\rm{LoS}}}} \right)\eta {P_u}} x_i^{ - {\alpha _d}}{g_i}G\left( {{\phi _i}} \right),
\end{aligned}
\end{equation}

\begin{equation}
\begin{aligned}
&{L_{I_u^c}}\left( {\frac{{\beta x_0^{{\alpha _u}}}}{{{P_u}G\left( {{\phi _0}} \right)}}} \right)\\& = {L_{I_{u,\rm{LoS}}^c}}\left( {\frac{{\beta x_0^{{\alpha _u}}}}{{{P_u}G\left( {{\phi _0}} \right)}}} \right){L_{I_{u,\rm{NLoS}}^c}}\left( {\frac{{\beta x_0^{{\alpha _u}}}}{{{P_u}G\left( {{\phi _0}} \right)}}} \right)\\
&= {E_{{\Phi _u}}}\left[ {\prod\limits_{{x_i} \in {\Phi _u}\backslash \{ 0\} } {\frac{1}{{1 + \beta x_0^{{\alpha _u}}{P_{\rm{LoS}}}x_i^{ - {\alpha _u}}\frac{{G\left( {{\phi _{_i}}} \right)}}{{G\left( {{\phi _0}} \right)}}}}} } \right]\times
\\&{E_{{\Phi _u}}}\left[ {\prod\limits_{{x_i} \in {\Phi _u}\backslash \{ 0\} } {\frac{1}{{1 + \beta x_0^{{\alpha _u}}(1 - {P_{\rm{LoS}}})\eta x_i^{ - {\alpha _u}}\frac{{G\left( {{\phi _{_i}}} \right)}}{{G\left( {{\phi _0}} \right)}}}}} } \right].
\end{aligned}
\end{equation}

${L_{I_{u,{\text{LoS}}}^c}}( {\frac{{\beta x_0^{{\alpha _u}}}}{{{P_u}}}} )$
and ${L_{I_{u,{\text{NLoS}}}^c}}( {\frac{{\beta x_0^{{\alpha _u}}}}{{{P_u}}}} )$ can be derived using the probability generation function
in (\ref{eq_2gaindulos}) and (\ref{eq_2gaindunlos}), where
${H_{11}}( {\beta ,{x_0},h,{\alpha _u}} )$ and ${H_{12}}( {\beta ,{x_0},h,{\alpha _u}} )$ are provided in (\ref{eq_h11}) and (\ref{eq_h12}), respectively.

\begin{figure*}[ht]
	\normalsize
\begin{equation}\label{eq_2gaind2dlos}
\begin{aligned}
{L_{{I_{u,\rm{LoS}}}}}( {\frac{{\beta d_0^{{\alpha _d}}}}{{{P_d}}}} )
& = \exp ( { - {\lambda _{2Du}}\int\limits_V {( {1 - \frac{1}{{1 + \frac{{\beta d_0^{{\alpha _d}}}}{{{P_d}}}{P_u}{P_{\rm{LoS}}}x_i^{ - {\alpha _u}}G( {{\varphi _i}} )}}} )} dx} )\\
& = \exp ( { - 2\pi {\lambda _{2Du}}\int_0^\infty  {( {1 - \frac{1}{{1 + \frac{{\beta d_0^{{\alpha _d}}}}{{{P_d}}}{P_u}{{(\sqrt {{r^2} + {h^2}} )}^{ - {\alpha _u}}}G( {\arctan (r/h)} )\frac{1}{{1 + C\exp ( { - B(\frac{{180}}{\pi }arctan(h/r) - C)} )}}}}} )rdr} } )\\
& = \exp ( { - 2\pi {\lambda _{2Du}}{H_9}( {\beta ,{d_0},h,{\alpha _d},{\alpha _u}} )} ),
\end{aligned}
\end{equation}

\begin{equation}\label{eq_2gaind2dnlos}
\begin{aligned}
{L_{{I_{u,\rm{NLoS}}}}}( {\frac{{\beta d_0^{{\alpha _d}}}}{{{P_d}}}} ) = \exp ( { - 2\pi {\lambda _{2Du}}{H_{10}}( {\beta ,{d_0},h,{\alpha _d},{\alpha _u}} )} ),
\end{aligned}
\end{equation}
where
\begin{equation}\label{eq_h9}
\begin{aligned}
{H_{9}}( {\beta ,{d_0},h,{\alpha _d},{\alpha _u}} ) = \int_0^\infty  {( {1 - \frac{1}{{1 + \frac{{\beta d_0^{{\alpha _d}}}}{{{P_d}}}{P_u}{{(\sqrt {{r^2} + {h^2}} )}^{ - {\alpha _u}}}G( {\arctan (r/h)} )\frac{1}{{1 + C\exp ( { - B(\frac{{180}}{\pi }arctan(h/r) - C)} )}}}}} )rdr},
\end{aligned}
\end{equation}

\begin{equation}\label{eq_h10}
\begin{aligned}
{H_{10}}( {\beta ,{d_0},h,{\alpha _d},{\alpha _u}} ){=}\int_0^\infty  {( {1 - \frac{1}{{1 + \frac{{\beta d_0^{{\alpha _d}}}}{{{P_d}}}{P_u}{{(\sqrt {{r^2} + {h^2}} )}^{ - {\alpha _u}}}\eta G( {\arctan (r/h)} ){(1{-}\frac{1}{{1 + C\exp ( { - B(\frac{{180}}{\pi }arctan(h/r) - C)} )}})}}}} )rdr}.
\end{aligned}
\end{equation}

\begin{equation}\label{eq_2gaindulos}
\begin{aligned}
{L_{I_{u,\rm{LoS}}^c}}( {\frac{{\beta x_0^{{\alpha _u}}}}{{{P_u}G( {{\phi _0}} )}}} )
&= \exp ( { - {\lambda _{2Du}}\int\limits_V {( {1 - \frac{1}{{1 + \beta x_0^{{\alpha _{u}}}{P_{\rm{LoS}}}x_i^{ - {\alpha _u}}\frac{{G( {{\phi _{_i}}} )}}{{G( {{\phi _0}} )}}}}} )} dx} )\\
&= \exp ( { - 2\pi {\lambda _{2Du}}\int_0^\infty  {( {1 - \frac{1}{{1 + \beta x_0^{{\alpha _u}}{{(\sqrt {{r^2} + {h^2}} )}^{ - {\alpha _u}}}\frac{{G( {\arctan (r/h)} )}}{{G( {{\phi _0}} )}}\frac{1}{{1 + C\exp ( { - B(\frac{{180}}{\pi }arctan(h/r) - C)} )}}}}} )rdr} } )\\
&= \exp ( { - 2\pi {\lambda _{2Du}}{H_{11}}( {\beta ,{x_0},h,{\alpha _u}} )} ),
\end{aligned}
\end{equation}

\begin{equation}\label{eq_2gaindunlos}
\begin{aligned}
{L_{I_{u,{\rm{NLoS}}}^c}}( {\frac{{\beta x_0^{{\alpha _u}}}}{{{P_u}G( {{\phi _0}} )}}} ) = \exp ( { - 2\pi {\lambda _{2Du}}{H_{12}}( {\beta ,{x_0},h,{\alpha _u}} )} ),
\end{aligned}
\end{equation}
where
\begin{equation}\label{eq_h11}
\begin{aligned}
{H_{11}}( {\beta ,{x_0},h,{\alpha _u}} ) = \int_0^\infty  {( {1 - \frac{1}{{1 + \beta x_0^{{\alpha _u}}{{(\sqrt {{r^2} + {h^2}} )}^{ - {\alpha _u}}}\frac{{G( {\arctan (r/h)} )}}{{G( {{\phi _0}} )}}\frac{1}{{1 + C\exp ( { - B(\frac{{180}}{\pi }arctan(h/r) - C)} )}}}}} ){r}dr},
\end{aligned}
\end{equation}

\begin{equation}\label{eq_h12}
\begin{aligned}
{H_{12}}( {\beta ,{x_0},h,{\alpha _u}} ){=}\int_0^\infty  {( {1 - \frac{1}{{1 + \beta x_0^{{\alpha _u}}{{(\sqrt {{r^2} + {h^2}} )}^{ - {\alpha _u}}}\eta \frac{{G( {\arctan (r/h)} )}}{{G( {{\phi _0}} )}}{(1{-}\frac{1}{{1 + C\exp ( { - B(\frac{{180}}{\pi }arctan(h/r) - C)} )}})}}}} )rdr}.
\end{aligned}
\end{equation}
\hrulefill
\end{figure*}

\subsubsection{Spectrum sharing of 3D UAV network}

\begin{figure*}[ht]
	\normalsize
\begin{equation}\label{eq_3gaind2dlos}
{L_{{I_{u,\rm{LoS}}}}}( {\frac{{\beta d_0^{{\alpha _d}}}}{{{P_d}}}}) = \exp ( { - {\lambda _{3Du}}\int\limits_V {( {1 - \frac{1}{{1 + \frac{{\beta d_0^{{\alpha _d}}}}{{{P_d}}}{P_u}{P_{\rm{LoS}}}x_i^{ - {\alpha _u}}G( {{\varphi _i}} )}}} )} dx} )= \exp ( { - 2\pi {\lambda _{3Du}}{H_{13}}( {\beta ,{d_0},h,{\alpha _d},{\alpha _u}} )} ),
\end{equation}

\begin{equation}\label{eq_3gaind2dnlos}
{L_{{I_{u,\rm{NLoS}}}}}\left( {\frac{{\beta d_0^{{\alpha _d}}}}{{{P_d}}}} \right) = \exp \left( { - 2\pi {\lambda _{3Du}}{H_{14}}\left( {\beta ,{d_0},h,{\alpha _d},{\alpha _u}} \right)} \right),
\end{equation}
where
\begin{equation}\label{eq_h13}
\begin{aligned}
{H_{13}}( {\beta ,{d_0},h,{\alpha _d},{\alpha _u}} ) = \int_{h_1}^{h_2} {\int_0^\infty  {( {1 - \frac{1}{{1 + \frac{{\beta d_0^{{\alpha _d}}}}{{{P_d}}}{P_u}{{(\sqrt {{r^2} + {z^2}} )}^{ - {\alpha _u}}}G( {\arctan (r/z)} )\frac{1}{{1 + C\exp ( { - B(\frac{{180}}{\pi }arctan(z/r) - C)} )}}}}} )} } rdrdz,
\end{aligned}
\end{equation}

\begin{equation}\label{eq_h14}
\begin{aligned}
{H_{14}}( {\beta ,{d_0},h,{\alpha _d},{\alpha _u}} ) = \int_{h_1}^{h_2} {\int_0^\infty  {( {1 - \frac{1}{{1 + \frac{{\beta d_0^{{\alpha _d}}}}{{{P_d}}}{P_u}\eta {{(\sqrt {{r^2} + {z^2}} )}^{ - {\alpha _u}}}G( {\arctan (r/z)} )(1 - \frac{1}{{1 + C\exp ( { - B(\frac{{180}}{\pi }arctan(z/r) - C)} )}})}}} )} } rdrdz.
\end{aligned}
\end{equation}

\begin{equation}\label{eq_3gaindulos}
{L_{I_{u,\rm{LoS}}^c}}( {\frac{{\beta x_0^{{\alpha _u}}}}{{{P_u}G( {{\phi _0}} )}}} ) = \exp ( { - {\lambda _{3Du}}\int\limits_V {( {1 - \frac{1}{{1 + \beta x_0^{{\alpha _{u}}}{P_{\rm{LoS}}}x_i^{ - {\alpha _u}}\frac{{G( {{\phi _i}} )}}{{G( {{\phi _0}} )}}}}} )} dx} )= \exp ( { - 2\pi {\lambda _{3Du}}{H_{15}}( {\beta ,{x_0},h,{\alpha _u}} )}),
\end{equation}

\begin{equation}\label{eq_3gaindunlos}
\begin{aligned}
{L_{I_{u,{\rm{NLoS}}}^c}}( {\frac{{\beta x_0^{{\alpha _u}}}}{{{P_u}}}} ) = \exp ( { - 2\pi {\lambda _{3Du}}{H_{16}}( {\beta ,{x_0},h,{\alpha _u}} )} ),
\end{aligned}
\end{equation}
where
\begin{equation}\label{eq_h15}
\begin{aligned}
{H_{15}}( {\beta ,{x_0},h,{\alpha _u}} ) = \int_{h1}^{h2} {\int_0^\infty  {( {1 - \frac{1}{{1 + \beta x_0^{{\alpha _u}}{{(\sqrt {{r^2} + {z^2}} )}^{ - {\alpha _u}}}\frac{{G( {\arctan (r/z)} )}}{{G( {{\phi _0}} )}}\frac{1}{{1 + C\exp ( { - B(\frac{{180}}{\pi }arctan(z/r) - C)} )}}}}} )} } rdrdz,
\end{aligned}
\end{equation}

\begin{equation}\label{eq_h16}
\begin{aligned}
{H_{16}}( {\beta ,{x_0},h,{\alpha _u}} ){=}\int_{h1}^{h2} {\int_0^\infty  {( {1 - \frac{1}{{1 + \beta x_0^{{\alpha _u}}\eta {{(\sqrt {{r^2} + {z^2}} )}^{ - {\alpha _u}}}\frac{{G( {\arctan (r/z)} )}}{{G( {{\phi _0}} )}}(1 - \frac{1}{{1 + C\exp ( { - B(\frac{{180}}{\pi }arctan(z/r) - C)} )}})}}} )} } rdrdz.
\end{aligned}
\end{equation}
\hrulefill
\end{figure*}

For ground network user,
${L_{I_{u,{\text{LoS}}}}}( {\frac{{\beta d_0^{{\alpha _d}}}}{{{P_d}}}} )$
and ${L_{I_{u,{\text{NLoS}}}}}( {\frac{{\beta d_0^{{\alpha _d}}}}{{{P_d}}}} )$ can be obtained using the probability generation function
in (\ref{eq_3gaind2dlos}) and (\ref{eq_3gaind2dnlos}), where
${H_{13}}( {\beta ,{d_0},h,{\alpha _u}} )$ and ${H_{14}}( {\beta ,{d_0},h,{\alpha _u}} )$ are provided in (\ref{eq_h13}) and (\ref{eq_h14}), respectively.

For UAV network user,
${L_{I_{u,{\text{LoS}}}^c}}( {\frac{{\beta x_0^{{\alpha _u}}}}{{{P_u}}}} )$
and ${L_{I_{u,{\text{NLoS}}}^c}}( {\frac{{\beta x_0^{{\alpha _u}}}}{{{P_u}}}} )$ can be derived using the probability generation function
in (\ref{eq_3gaindulos}) and (\ref{eq_3gaindunlos}), where
${H_{15}}( {\beta ,{x_0},h,{\alpha _u}} )$ and ${H_{16}}( {\beta ,{x_0},h,{\alpha _u}} )$ are derived in (\ref{eq_h15}) and (\ref{eq_h16}).

\section{Simulation Results and Analysis}

In this section,
the simulation results regarding the
performance of UAV network and ground network with spectrum sharing are provided.
The main parameters for simulation are summarized in Table \ref{tab_2}.

\begin{table}[!t]
	\caption{Simulation parameters}
    \label{tab_2}
	\begin{center}
		\begin{tabular}{l|l}
			\hline
			\hline
			{Parameter} & {Value} \\
			\hline
			$P_u$ & 5 W \\
			$P_d$ & 0.1 W \\
			$\alpha_u$ & 3 \\
			$\alpha_d$ & 4 \\
			$B$ and $C$ & 0.136 and 11.95\\
			$\beta$ & 0.1 \\
			$\eta$ & 0.001\\
			$\lambda_u$ & $10^{-4}$ per square meter\\
			$\lambda_d$ & $10^{-3}$ per square meter\\
			$d_0$ & 10 m\\
			$N$ & $10^{-9}$ W\\
			\hline
			\hline
		\end{tabular}
	\end{center}
\end{table}

\subsection{The performance of ground network user}
\subsubsection{Omnidirectional A2A transmission}

\begin{figure}[!t]
	\centering
	\includegraphics[width=0.48\textwidth]{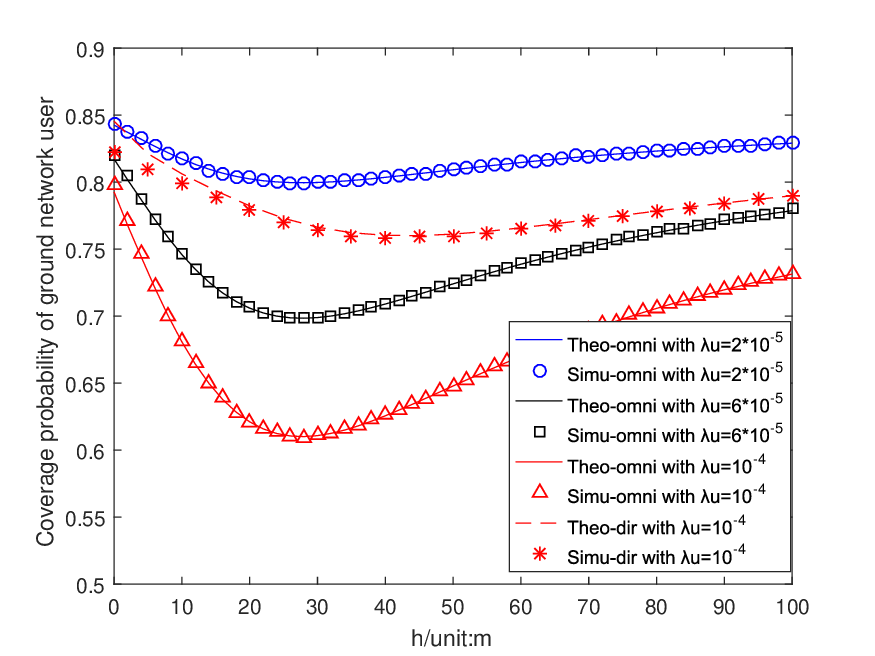}
	\caption{The coverage probability
		of ground network user versus $h$
		for 2D UAV deployment and omnidirectional A2A transmission, where ``theo/simu-omni/dir'' refers to  theoretical/simulation results
of omni/directional A2A transmission.}
	\label{Figure_4}
\end{figure}

\begin{figure}[!t]
	\centering
	\includegraphics[width=0.48\textwidth]{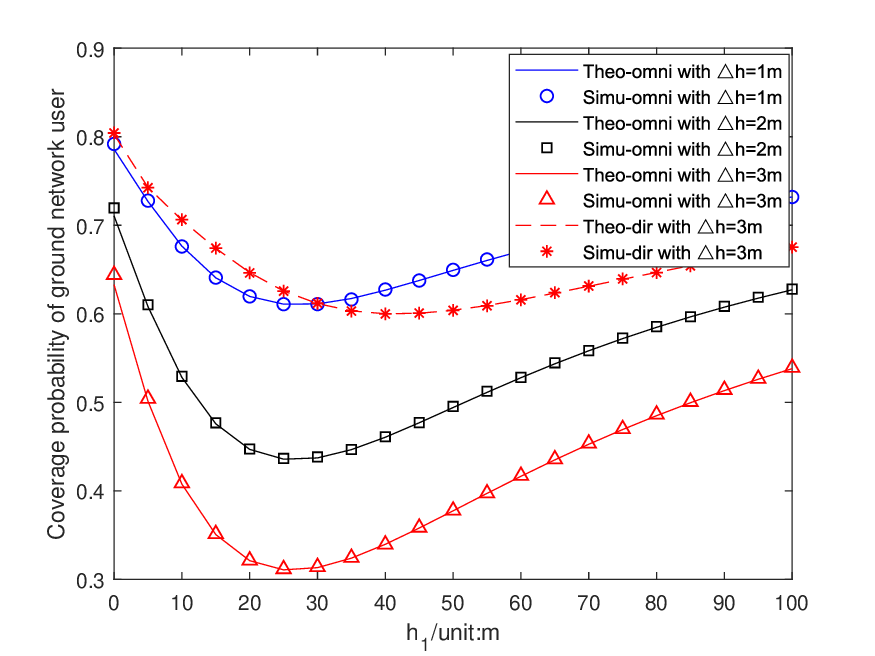}
	\caption{The coverage probability
		of ground network user versus $h_1$
		for 3D UAV deployment and omnidirectional A2A transmission, where ``theo/simu-omni/dir''
refers to theoretical/simulation results
of omni/directional A2A transmission.}
	\label{Figure_5}
\end{figure}

$P_1$ is illustrated in
Fig. \ref{Figure_4} and Fig. \ref{Figure_5}.
The Monte Carlo results, each of which runs $10^6$ times simulations, are
provided.
The solid lines denoting the theoretical results fit well with the points denoting the Monte Carlo results.
In Fig. \ref{Figure_4},
$P_1$ is a unimodal function of $h$.
When $h$ is smaller than a threshold,
the increase of $h$ will result in the increase of the probability of LoS transmission, which will decrease $P_1$.
However, when $h$ is larger than a threshold,
the increase of $h$ will enlarge
the distance between UAVs and the typical user of ground network,
which will
increase $P_1$.
In Fig. \ref{Figure_4}, with the increase of
$\lambda_u$, the interference from UAVs to the typical user of ground network increases. As a result, the coverage probability of ground network user decreases in this situation.

Fig. \ref{Figure_5} illustrates $P_1$
versus $h_1$ with 3D UAV deployment.
$P_1$ is also a unimodal function of $h$.
On the other hand,
$P_1$ is a decreasing function of $\Delta h$.

\subsubsection{Directional A2A transmission}

\begin{figure}[!t]
	\centering
	\includegraphics[width=0.48\textwidth]{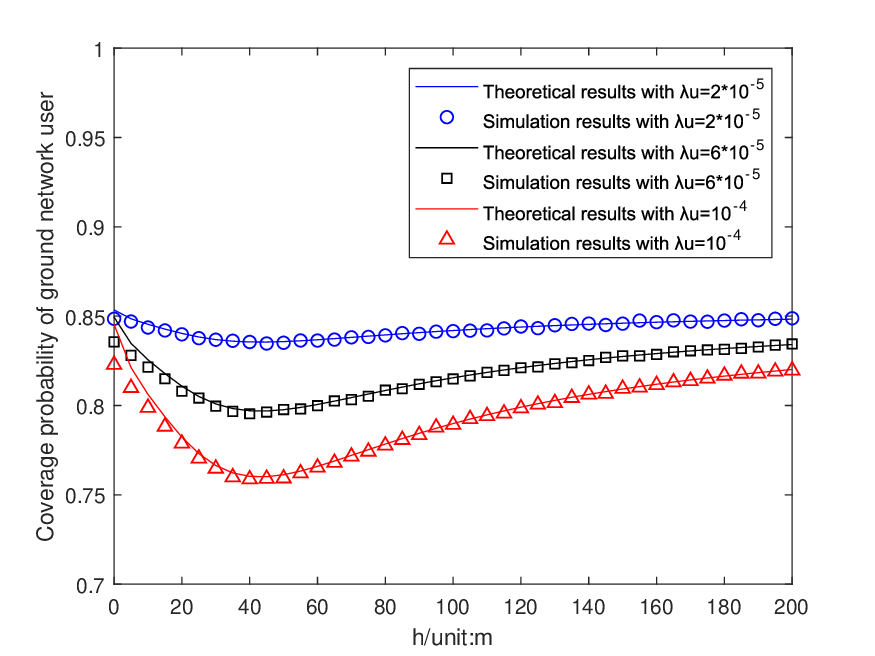}
	\caption{The coverage probability of ground
		network user versus $h$
		for 2D UAV deployment and directional A2A transmission.}
	\label{Figure_6}
\end{figure}

\begin{figure}[!t]
	\centering
	\includegraphics[width=0.48\textwidth]{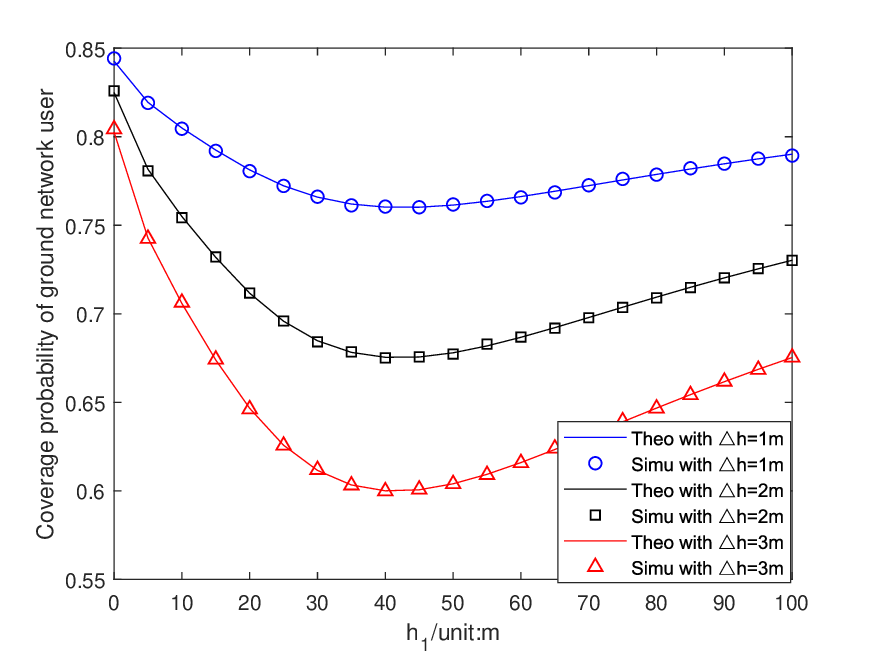}
	\caption{The coverage probability of ground network user versus $h_1$ for 3D UAV deployment and directional A2A transmission, where ``theo/simu'' refers to theoretical/simulation results.}
	\label{Figure_7}
\end{figure}

$P_1$ versus height with
2D and 3D UAV deployment is illustrated in
Fig. \ref{Figure_6} and Fig. \ref{Figure_7}.
Comparing Fig. \ref{Figure_6} and Fig. \ref{Figure_7}
with Fig. \ref{Figure_4} and Fig. \ref{Figure_5}, respectively,
we discover that $P_1$ with directional transmission
is larger than that with omnidirectional transmission.
According to Fig. \ref{Figure_4} and
Fig. \ref{Figure_5},
the coverage probability of the typical user in ground network
with directional transmission
is larger than that with omnidirectional transmission.

\subsection{The performance of UAV network user}

\subsubsection{Omnidirectional A2G transmission}

\begin{figure}[!t]
	\centering
	\includegraphics[width=0.48\textwidth]{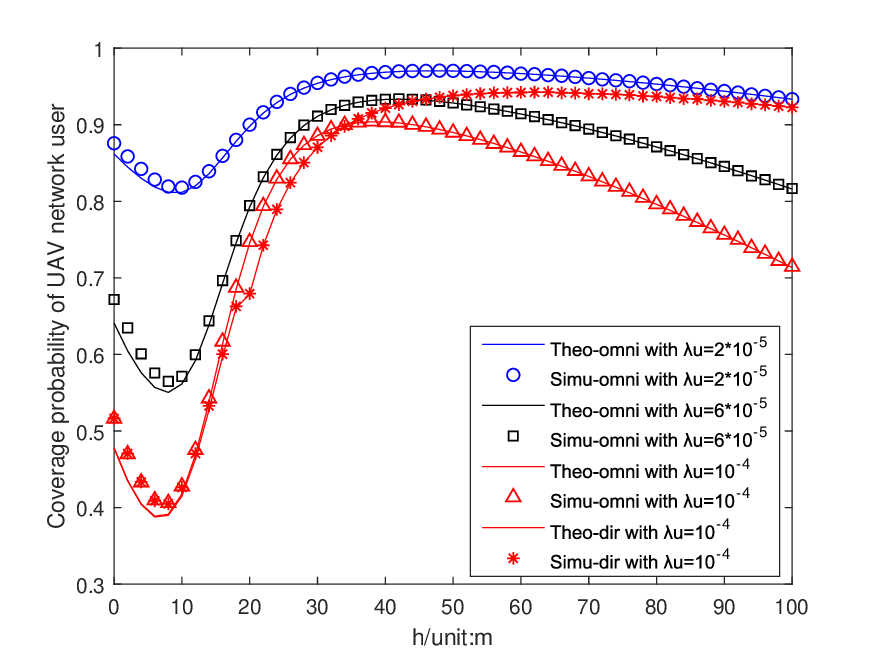}
	\caption{The coverage probability
		of UAV network user versus $h$
		for 2D UAV deployment and omnidirectional A2A transmission, where ``theo/simu-omni/dir''
refers to theoretical/simulation results
of omni/directional A2A transmission.}
	\label{Figure_8}
\end{figure}

\begin{figure}[!t]
	\centering
	\includegraphics[width=0.48\textwidth]{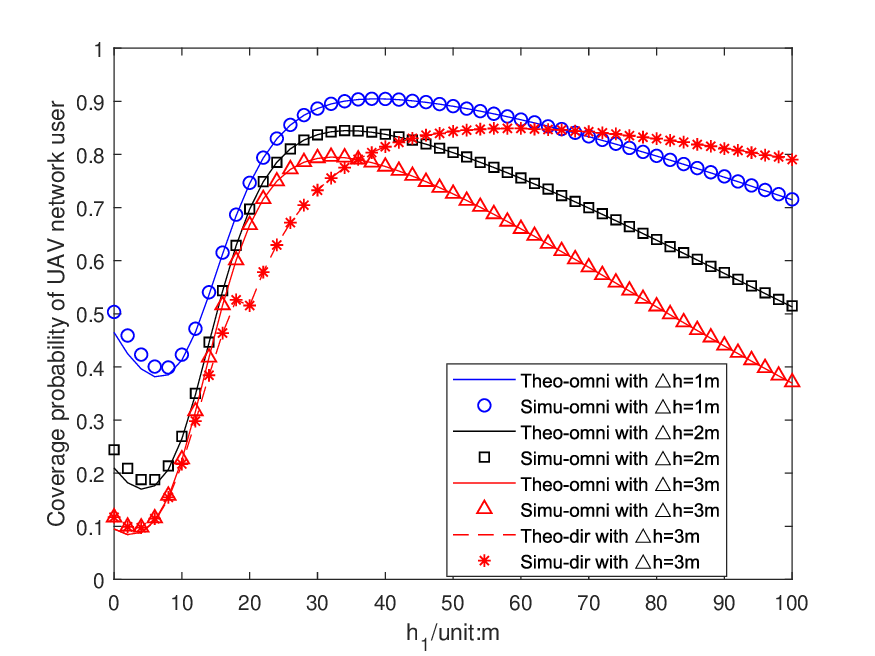}
	\caption{The coverage probability of
		UAV network user versus $h_1$
		for 3D UAV deployment and omnidirectional A2A transmission, where ``theo/simu-omni/dir'' refers to theoretical/simulation results
of omni/directional A2A transmission.}
	\label{Figure_9}
\end{figure}

$P_2$ is illustrated
in Fig. \ref{Figure_8} and Fig. \ref{Figure_9}.
The Monte Carlo results, each of which
runs $10^6$ times simulations, are provided.
The lines denoting the theoretical results fit well with the dots denoting Monte Carlo results.
For 2D and 3D UAV networks,
there exists optimal height
to maximize $P_2$.
When UAVs are 2D deployed, $P_2$ is inversely proportional to the increase of UAVs’ density because users of typical UAV network receive increasing interference.
When UAVs are 3D deployed, $P_2$ is proportional to the increase of $\Delta h$ because the interference received by the typical UAV network user from interfering UAVs will decrease.

\subsubsection{Directional A2G transmission}

\begin{figure}[!t]
	\centering
	\includegraphics[width=0.48\textwidth]{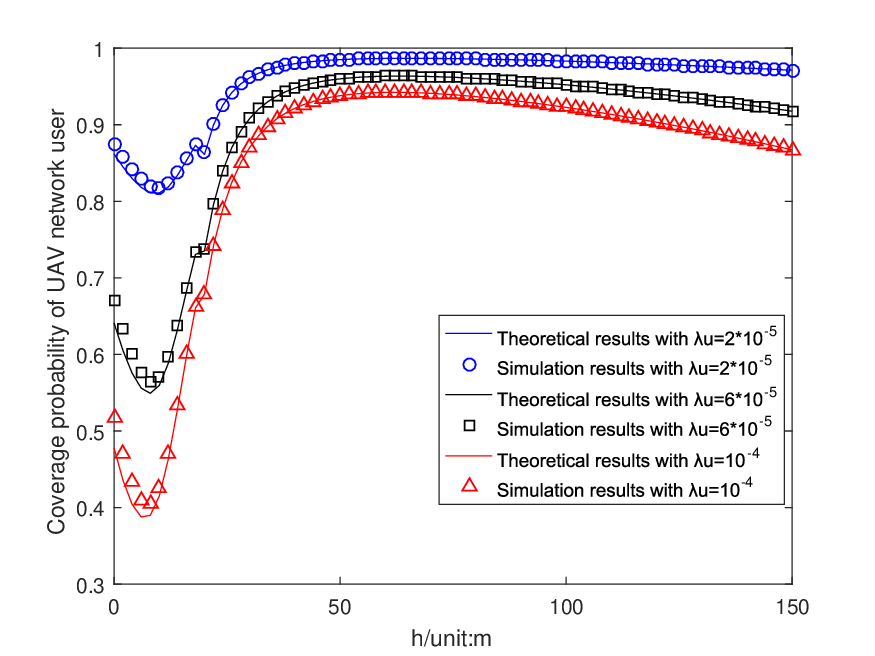}
	\caption{The coverage
		probability of UAV network user versus $h$
		for 2D UAV network and directional transmission.}
	\label{Figure_10}
\end{figure}

\begin{figure}[!t]
	\centering
	\includegraphics[width=0.48\textwidth]{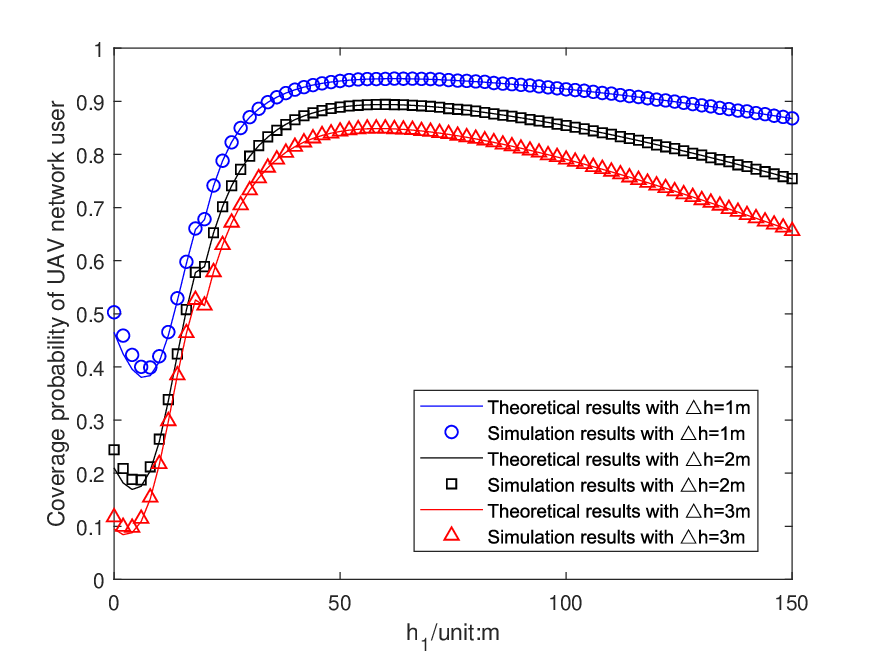}
	\caption{The coverage probability of UAV network user versus $h_1$
		for 3D UAV network and directional transmission.}
	\label{Figure_11}
\end{figure}

With directional A2G communication,
$P_2$ can be improved.
Comparing Fig. \ref{Figure_8} with Fig. \ref{Figure_10},
it can be found that for 2D UAV deployment,
$P_2$
with directional A2G transmission
is larger than that with omnidirectional A2G transmission,
especially in the situation with large heights.
The essential reason is that when a UAV is high above ground and adopts
omnidirectional A2G transmission,
UAV can generate interference to a large number of UAV network users.
However, with directional A2G transmission,
UAV will generate interference to a relatively limited number of UAV network users.
Similarly, comparing Fig. \ref{Figure_9} with Fig. \ref{Figure_11},
it can be discovered that for 3D UAV network,
$P_2$
with directional A2G transmission
is still larger than that with
omnidirectional A2G transmission especially when $h_1$ is large.

\subsection{Optimal height of UAVs}

The concept of
transmission capacity (TC) is adopted as the performance metric of UAV network,
which is as follows \cite{cog_zhangwei}.

\begin{equation}
T_u = {\lambda _u}P(\gamma_{uu} > \beta )\log (1 + \beta ),
\end{equation}
where $\gamma_{uu}$ denotes SINR. In this paper, transmission capacity ${T_u}$ refers to the sum capacity of UAV network users per unit area which can successful communication.
Solving the
optimization model as follows, the optimal altitude of UAVs
could be obtained.
\begin{equation}\label{eq_optimization}
\begin{aligned}
& \mathop {\max }\limits_{{h}} {\kern 3pt} {T_u}\\
& s.t.{\kern 9pt} {P_1} \ge \alpha.
\end{aligned}
\end{equation}

Since the inequalities in (\ref{eq_optimization}) is complex, the closed-form solution is hard to be obtained.
The optimal solution can be searched numerically.

\begin{figure}[!t]
	\centering
	\includegraphics[width=0.48\textwidth]{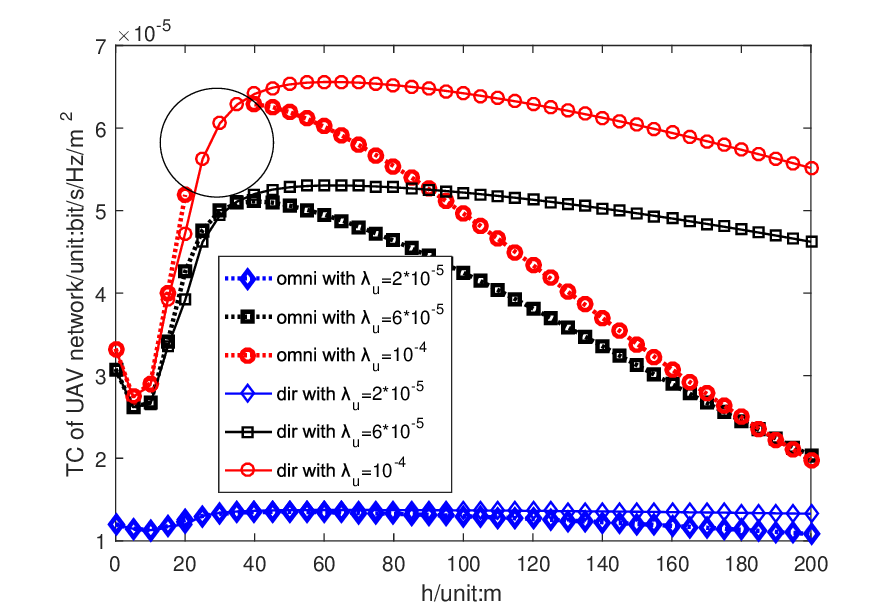}
	\caption{The transmission capacity
		of UAV network user versus $h$ for 2D UAV network, where ``omni/dir'' refers to omnidirectional/directional
A2A transmission.}
	\label{Figure_12}
\end{figure}

TC of 2D UAV network is illustrated in Fig. \ref{Figure_12}.
TC of UAV network is an increasing function of the density of UAVs.
Besides, the directional transmission can improve
TC of UAV network. Considering TC of UAV network changing with
h, there exists an optimal $h$ to maximize the TC of UAV network.
However, when $\lambda_u$ is highest in simulation,
for example, $\lambda_u = 10^{-4}$ and
omnidirectional transmission is adopted,
there is a vacant interval of $h$,
where the interference received by the ground network user from UAVs
is large and the constraint in (\ref{eq_optimization})
is not satisfied.
Hence, there exists a vacant interval in the curve,
which is denoted
by a circle in Fig \ref{Figure_12}.

\begin{figure}[!t]
	\centering
	\includegraphics[width=0.48\textwidth]{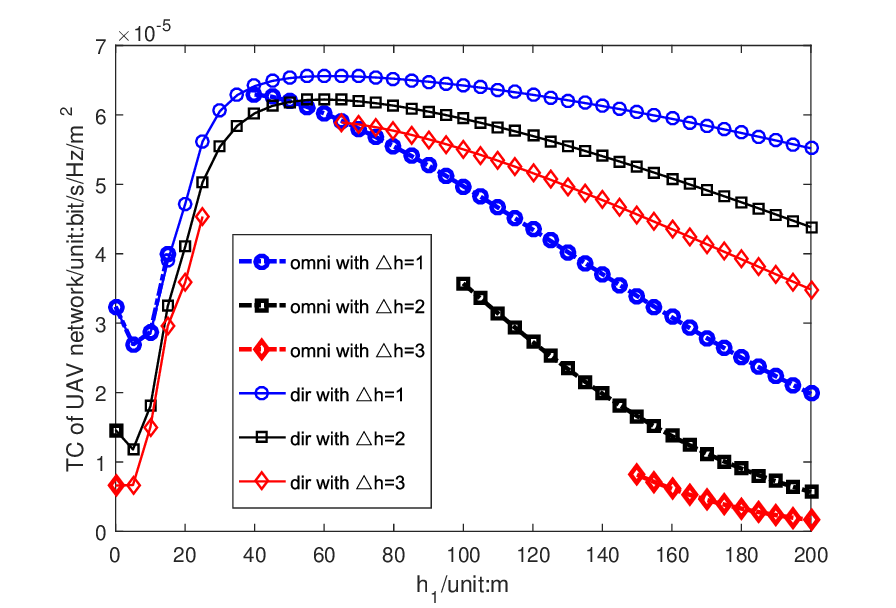}
	\caption{The transmission capacity of
		UAV network user versus $\Delta h$ for 3D UAV network, where ``omni/dir'' refers to omnidirectional/directional
A2A transmission.}
	\label{Figure_13}
\end{figure}

In Fig. \ref{Figure_13}, TC of 3D UAV network is decreasing with the increase of $\Delta h$.
Meanwhile, the directional transmission improves
TC of UAV network.
Similar to 2D UAV network,
there exist vacant intervals
for the curves of TC of 3D UAV network,
where the
interference received by ground network user from UAVs is
small and the constraint of (\ref{eq_optimization})
is not satisfied.

\section{Conclusion}

In this paper,
	the spectrum sharing between UAV enabled wireless mesh
	networks and ground networks is analyzed
	using stochastic geometry.
The correctness of theoretical derivation is verified by simulation results.
When $\Delta h \to 0$,
the derived results of 3D UAV network reduce
to that of 2D UAV network.
This is also verified in simulation results.
Compared with omnidirectional transmission,
the performance of spectrum sharing with
directional transmission is better.
We can find the optimal height of UAVs achieved to
	maximize the transmission capacity
	within the constraint
	of the coverage probability of ground network user.
This paper provides performance
analysis of
UAV network, which
may motivate the study of architecture, protocol and resource allocation for aerial wireless mesh networks
with spectrum sharing.

\end{document}